%% file: main.tex
\documentclass[runningheads]{llncs}

\usepackage{eccv}
\input{preamble}

\usepackage{eccvabbrv}
\usepackage{graphicx}
\usepackage{booktabs}
\usepackage[accsupp]{axessibility} 
\usepackage{hyperref}
\usepackage{orcidlink}
\usepackage{appendix}

\begin{document}

\def\OURS{Stable Video Portraits}
\title{Stable Video Portraits} 

\author{Mirela Ostrek\inst{1}\orcidlink{0009-0009-9987-646X} \and
Justus Thies\inst{1,2}\orcidlink{0000-0002-0056-9825}}

\authorrunning{M. Ostrek et al.}

\institute{Max Planck Institute for Intelligent Systems, Tübingen, Germany \and
Technical University of Darmstadt, Darmstadt, Germany
\\
}

\maketitle

\input{fig/figtex}
\input{sec/0_abstract}

\input{sec/1_intro}
\input{sec/2_related_work}
\input{sec/3_method}
\input{sec/4_experiments}
\input{sec/5_conclusion}
\noindent \textbf{Acknowledgements:}
The authors thank
Tsvetelina Alexiadis, Claudia Gallatz and Asuka Bertler for data collection;
Tsvetelina Alexiadis, Tomasz Niewiadomski and Taylor McConnell for perceptual study;
Yue Gao, Nikita Drobyshev, Jalees Nehvi and Wojciech Zielonka for help with the baselines;
Michael J. Black and Yao Feng for discussions;
and Benjamin Pellkofer for IT support.
This work has received funding from the European Union’s Horizon 2020 research and innovation programme under the Marie Skłodowska Curie grant agreement No 860768 (CLIPE project).

\clearpage

\bibliographystyle{splncs04}
\bibliography{main}

\clearpage
\input{sec/X_suppl}
\end{document}

%% file: preamble.tex
\usepackage[dvipsnames]{xcolor}

\usepackage{graphicx}

\usepackage{multirow}

\usepackage{array}

\usepackage{colortbl}

\usepackage{wrapfig}

\usepackage{tabularx}

\newcommand{\expec}{\mathbb{E}}

\newcommand{\lsimple}{L_{DM}}
\newcommand{\lsimpleldm}{L_{LDM}}

\newcommand{\model}{\epsilon_\theta}

\newcommand{\encoder}{\mathcal{E}}

%% file: fig/figtex.tex
\newcommand{\teaserCaption}{
\textit{\OURS} is a high-fidelity hybrid 2D/3D person-specific monocular head avatar method; fine-grained control over the head pose and expression parameters is achieved via 3DMM conditions.
Leveraging the Stable Diffusion prior, generated faces may be morphed into celeb faces using text, without any fine-tuning at test time.
}

\begin{center}
    \centering
    \captionsetup{type=figure}
    \includegraphics[width=\textwidth]{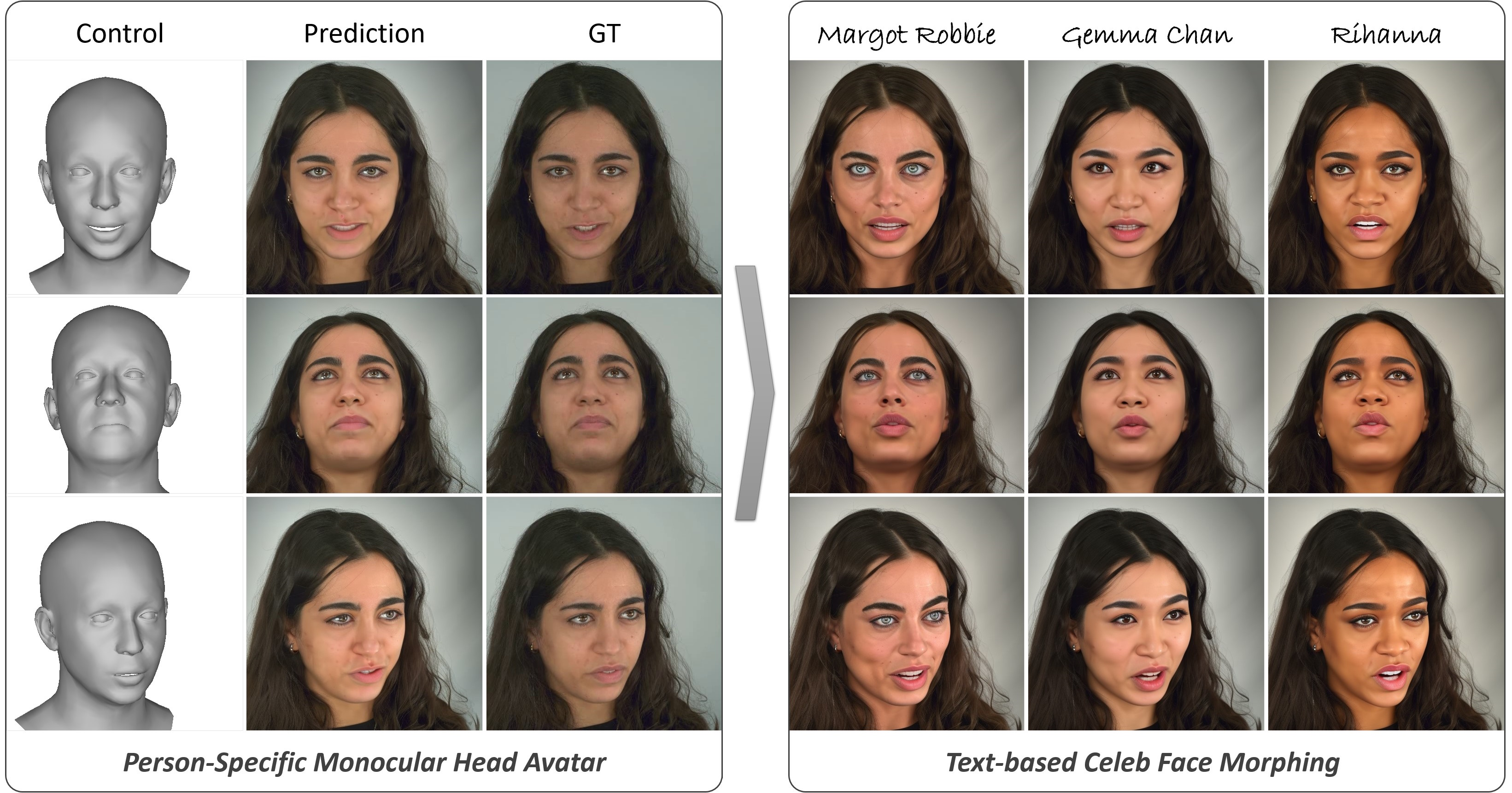}
    \caption{\teaserCaption}
    \label{fig:teaser}
\end{center}

%% file: sec/0_abstract.tex
\begin{abstract}
Rapid advances in the field of generative AI and text-to-image methods in particular have transformed the way we interact with and perceive computer-generated imagery today.
In parallel, much progress has been made in 3D face reconstruction, using 3D Morphable Models (3DMM). 
In this paper, we present \textit{\OURS}, a novel hybrid 2D/3D generation method that outputs photorealistic videos of talking faces leveraging a large pre-trained text-to-image prior (2D), controlled via a 3DMM (3D). 
Specifically, we introduce a person-specific fine-tuning of a general 2D stable diffusion model which we lift to a video model by providing temporal 3DMM sequences as conditioning and by introducing a temporal denoising procedure.
As an output, this model generates temporally smooth imagery of a person with 3DMM-based controls, i.e., a person-specific avatar.
The facial appearance of this person-specific avatar can be edited and morphed to text-defined celebrities, without any fine-tuning at test time.
The method is analyzed quantitatively and qualitatively, and we show that our method outperforms state-of-the-art monocular head avatar methods. 
\href{https://svp.is.tue.mpg.de/}{https://svp.is.tue.mpg.de/}
\keywords{Neural rendering \and Generative AI \and Head avatars}
\end{abstract}

%% file: sec/1_intro.tex
\section{Introduction}
\label{sec:intro}
Digitizing and animating our appearance is a core challenge for various human-centric applications including telepresence in AR or VR, e-commerce, or content creation for entertainment purposes (movies, computer games).
In the past years, we have seen immense progress in this field~\cite{zollhofer2018state,tewari2020state, tewari2022advances} ranging from classical 3D reconstruction methods that use 3DMMs~\cite{thies2016face2face}, point clouds~\cite{zheng2023pointavatar}, and meshes with dynamic offsets~\cite{grassal2022neural}, to 2D neural rendering~\cite{kim2018deep,chan2019everybody} and 3D neural rendering utilizing neural-radiance-based representations~\cite{gafni2021dynamic,zheng2022avatar}.
In parallel, we have witnessed tremendous successes in generative AI, especially, for image synthesis.
Most recently, large text-to-image latent diffusion models such as Stable Diffusion~\cite{stable_diffusion}, DALL-E~\cite{dalle, dalle-2}, and MidJourney~\cite{midjourney}, amongst others, have been in the spotlight due to their capacity to generate remarkably detailed images of high quality, in a matter of seconds.
To control the image generation process further, new finetuning modules and techniques such as ControlNet~\cite{controlnet}, LoRA~\cite{lora}, DreamBooth~\cite{dreambooth}, and textual inversion~\cite{textual_inversion1, textual_inversion2} have been proposed.
In this paper, we devise a hybrid 2D/3D monocular head avatar method called \textit{\OURS}, leveraging (i) a large pre-trained 2D text-to-image prior (Stable Diffusion) that is able to generate remarkably detailed, lifelike faces of high quality, and (ii) a 3DMM-based 3D head model that offers a high level of controllability of the avatar as it parameterizes the head in terms of 3D head pose and facial expressions~\cite{flame,bfm}.
To bridge the gap between 2D and 3D, the conditioning 3DMMs are projected onto a 2D image plane, and person-specific finetuning of stable diffusion via ControlNet~\cite{controlnet} is performed using a short video.
However, the process of generating 2D avatar images conditioned on the output of an off-the-shelf 3D face reconstruction method presents several challenges.
These challenges include temporal inconsistencies in reconstructed meshes and their corresponding 3DMM parameters, potential ambiguities in regions not controlled by the mesh (e.g., hair and shoulders), and discrepancies in texture and shape, attributable to inaccuracies in the underlying 3D model.
These issues often manifest as flickering artifacts in the resulting videos.
To address such challenges, we introduce novel techniques aimed at enhancing temporal consistency.
Firstly, temporal conditioning is incorporated into ControlNet by leveraging a sequence of meshes as input.
Secondly, the denoising process is refined by integrating the latent representation of the previous frame during inference.
These methodological improvements lead to predictions that exhibit greater stability and visual coherence throughout the generated videos.

In addition to generating person-specific avatars, our method allows for face morphing at test time to adjust the avatar's appearance to resemble a celebrity, guided by text.
During training, ControlNet learns consistent shape and texture representations.
At test time, it then generates consistent offsets to those representations to morph the person-specific face with the target celebrity face, leveraging existing celebrity appearances within the Stable Diffusion text-to-image prior. 
Remarkably, our method produces diverse appearances at test time from a single training sequence, eliminating the need for any finetuning or training for new text-defined identity morphs.

We demonstrate the effectiveness of our method through quantitative and qualitative ablation studies comparing it with state-of-the-art monocular head avatar methods.
Our method outperforms the state-of-the-art, especially considering its capability to reconstruct detailed images (LPIPS) of high-fidelity human faces (FID/KID).
\textbf{In summary}, we make the following \textit{contributions}:
\smallskip
\begin{itemize}
 \item[$\circ$] \textit{A hybrid 2D/3D monocular avatar method}, where (2D) Stable Diffusion image prior is leveraged to achieve high-fidelity reconstructions, while (3D) 3DMM enables fine-grained control over the head pose and expression parameters in a temporally coherent manner.
  \item[$\circ$] \textit{A novel denoising procedure}, where each video frame is generated considering the previous frame, at inference time, producing temporally stable avatars.
  \item[$\circ$] \textit{A technique to morph the facial appearance} of a person-specific avatar, into a text-defined celebrity, without any fine-tuning at test time.
  \item[$\circ$] \textit{A portrait avatar dataset} with long talking sequences (8+ minutes) containing examples of non-trivial head movements for 6 female participants. Please see our supplementary materials for more details.
\end{itemize}

%% file: sec/2_related_work.tex
\section{Related Work}
\label{sec:related}
Our proposed approach integrates aspects of 3DMM-based avatar reconstruction with 2D generative AI, forming a hybrid methodology for reconstructing portrait avatars from person-specific video sequences. 
In this section, we explore related work on facial avatar reconstruction from monocular data, and 2D generative methods
including the techniques leveraging large pre-trained diffusion models~\cite{stable_diffusion}, as well as Generative Adversarial Networks (GANs)~\cite{goodfellow2020generative}.

\paragraph{\textbf{3D Head Avatars:}}
In the past few years, we have seen a large amount of publications about 3D human avatars.
Especially, for facial avatars there are state-of-the-art reports that cover classical 3DMM-based methods~\cite{zollhofer2018state}, as well as 2D/3D neural rendering-based methods~\cite{tewari2020state,tewari2022advances}.
The majority of methods are based on a 3D morphable model like FLAME~\cite{flame} or BFM~\cite{bfm} to allow for expression and pose control.
NeRFace~\cite{gafni2021dynamic} embeds a neural radiance field around the BFM model which is conditioned on the expression codes of the morphable model which effectively leads to a dynamic neural radiance field.
In INSTA~\cite{zielonka2023instant}, the neural radiance field is deformed with the deformation field spanned by the morphable model.
I'M Avatar~\cite{zheng2022avatar} uses a signed distance function to model the surface of a human which is also driven by the deformation induced by the morphable model.
In contrast to these implicit representations, NHA~\cite{grassal2022neural} uses an explicit mesh representation, by learning pose and expression dependent displacements on top of the FLAME template mesh.
PointAvatar~\cite{zheng2023pointavatar} represents the geometry of a face using surface sample points. The resulting point cloud can be splatted, but holes can occur in the output animation.
DVP~\cite{kim2018deep} and DNR~\cite{thies2019deferred} combine 3DMM-based rendering with 2D image synthesis using Pix2Pix-like U-Net architectures~\cite{pix2pix2017} which is referred to as 2D neural rendering~\cite{tewari2022advances}.
Besides these 2D refinement methods, there are also warping-based methods that directly leverage the information of the training input during inference.
PECHead~\cite{gao2023high} is using landmarks of a 3DMM to warp an intermediate feature representation of a single input image to follow the motion of a driving video. In addition to the 3DMM landmarks it also leverages learned key-points similar to the first order motion models (FOMM)~\cite{fomm}.
MegaPortraits~\cite{drobyshev2022megaportraits} is a generalized single shot method, but it encodes an image into a 3D volumetric space which can be warped by a driving sequence and decoded to an output image.
In contrast to the above-mentioned methods that are based on commodity sensors such as single webcams, there are methods based on multi-view data that produce high-quality 3D avatars ('codec avatars'~\cite{ma2021pixel}, MVP~\cite{Lombardi21}).
Cao at al.~\cite{cao2022phone} demonstrate how to use high-quality data to train a generative model for 3D face appearances which can be fine-tuned for a specific subject using one monocular sequence.
Our method does not require a large-scale 3D dataset, instead, it uses the Stable Diffusion image prior (trained on a collection of 2D images).
\paragraph{\textbf{2D Generative AI:}}
StyleGAN-based methods~\cite{karras2019style, karras2020analyzing, karras2021alias}  are widely recognized for their ability to generate high-quality images of human faces using GANs~\cite{goodfellow2020generative}.
StyleAvatar~\cite{wang2023styleavatar} leverages the StyleGAN architecture to produce UV maps for a 3DMM-based neural avatar~\cite{thies2019deferred}.
In contrast, diffusion models~\cite{dalle,dalle-2,midjourney,stable_diffusion,imagen} have gained prominence more recently, surpassing previous image synthesis techniques in terms of output quality and diversity.
These large-scale models are trained on extensive image datasets such as LAION-5B~\cite{schuhmann2022laion} that include textual descriptions, facilitating text-to-image generation.
Text conditioning is important for our method as it allows for the modification of the identity of the person-specific avatar.
We use Stable Diffusion (LDM)~\cite{stable_diffusion} as the image prior and further condition the model using ControlNet ~\cite{controlnet}.
The latter facilitates image-to-image translation by incorporating input conditioning, such as 2D images (e.g., landmarks, segmentation masks), into latent diffusion models.

\paragraph{\textbf{Video Diffusion:}}
Similarly, significant progress has been made in the field of video-to-video translation and related domains~\cite{yang2023rerender, geyer2023tokenflow, ceylan2023pix2video, qi2023fatezero, wu2023tune}.
In contrast to these works, our method excels in generating unseen sequences of subjects at test time, using a sequence of 3DMMs to enable modifications of shape, pose, and expressions in 3D.
Additionally, our approach empowers the alteration of the original identity's appearance, through morphing with the face of a celebrity using text alone and without requiring test-time finetuning.

Recently, many text-to-video models such as Lumiere~\cite{bar2024lumiere}, SORA~\cite{videoworldsimulators2024}, and Stable Video Diffusion~\cite{blattmann2023stable} have appeared. 
Some of these models excel in generating coherent sequences from images or textual descriptions but struggle with maintaining high quality over extended durations and require significant computational resources. 
In contrast, single-frame generation models allow for unlimited frame generation without quality degradation and offer greater control, customization, and resource efficiency. 
Here, we explore the time-space consistency of single-frame video models.

%% file: sec/3_method.tex
\section{Background}
\OURS~ is built using the Stable Diffusion image prior~\cite{stable_diffusion} and the ControlNet mechanism~\cite{controlnet}.
In the following, the related notation is introduced.
\paragraph{\textbf{Denoising diffusion probabilistic model (DDPM):}}
A denoising diffusion probabilistic model is trained to generate samples of a data distribution via an interative denoising procedure that starts at a sample of pure Gaussian noise.
It is trained to denoise synthetically corrupted data samples.
Specifically, given a sample image $\mathbf{x}_0$  drawn from the original data distribution $q(\mathbf{x})$; $\mathbf{x}_0 \sim q(\mathbf{x})$, Gaussian noise $\mathcal{N}(\mathbf{0}, \mathbf{I})$ is gradually added to $\mathbf{x}_0$ in $T$ steps (Markov process), as defined by a variance schedule $\beta_t$, where $\{\beta_t \in (0, 1)\}_{t=1}^T$.
This is a forward diffusion process, where each step is defined as:
\begin{equation}
    \begin{split}
    q(\mathbf{x}_t \vert \mathbf{x}_{t-1}) = \mathcal{N}(\mathbf{x}_t; \sqrt{1 - \beta_t} \mathbf{x}_{t-1}, \beta_t\mathbf{I}) .
    \end{split}
\end{equation}
At the end of the degradation process ($t=T$), $\mathbf{x}_0$ is reduced to pure Gaussian noise.
For the duration of all of the time-steps $t$ of this forward diffusion process, a neural network is trained to gradually denoise the sample $\mathbf{x}_t$ and retrieve the less noisy sample $\mathbf{x}_{t-1}$:
\begin{equation}
    \begin{split}
    p_\theta(\mathbf{x}_{t-1} \vert \mathbf{x}_t) = \mathcal{N}(\mathbf{x}_{t-1}; \boldsymbol{\mu}_\theta(\mathbf{x}_t, t), \boldsymbol{\Sigma}_\theta(\mathbf{x}_t, t)).
    \end{split}
\end{equation}
In our case, the denoising network with parameters $\theta$ is a U-Net~\cite{pix2pix2017} and it is trained using the objective~\cite{ho2020denoising}:
\begin{equation}
\lsimple = \expec_{x, \epsilon \sim \mathcal{N}(0, 1),  t }\Big[ \Vert \epsilon - \model(x_{t},t) \Vert_{2}^{2}\Big] \, ,
\label{eq:dmloss}
\end{equation}
where noise $\epsilon$ is predicted instead of the denoised sample.
Using this trained denoising network, new images can be generated by reverting the entire forward diffusion process, staring at pure noise $\mathbf{x}_T \sim \mathcal{N}(\mathbf{0}, \mathbf{I})$:
\begin{equation}
\begin{split}
p_\theta(\mathbf{x}_{0:T}) = p(\mathbf{x}_T) \prod^T_{t=1} p_\theta(\mathbf{x}_{t-1} \vert \mathbf{x}_t) .
\end{split}
\label{eq:denoising}
\end{equation}
For a time-efficient inference procedure that avoids using $T$ denoising steps, denoising diffusion implicit models (DDIM)~\cite{song2020denoising} were introduced with a modified sampling schedule (at test time), where only a small subset of $S$ DDPM diffusion steps
is utilized to predict the final output.

\paragraph{\textbf{Latent diffusion model (LDM):}}
Instead of applying the diffusion model directly in image space, it can be applied on a low-dimensional latent representation of an image, thus, improving the training and inference efficiency~\cite{stable_diffusion}.
Encoding a pixel space sample $\mathbf{x}_{t}$ into a latent space sample $\mathbf{z}_{t}$ is achieved via an encoder $\encoder$ from a variational auto-encoder (VAE), resulting in:
\begin{equation}
\lsimpleldm := \expec_{\encoder(x), \epsilon \sim \mathcal{N}(0, 1),  t}\Big[ \Vert \epsilon - \model(z_{t},t) \Vert_{2}^{2}\Big] \, .
\label{eq:ldmloss}
\end{equation}
Using the decoder of the VAE, denoised latent samples can be converted back to the actual images~\cite{kingma2013auto}.
\paragraph{\textbf{Conditional LDM:}}
An important aspect of the diffusion models is that the denoising procedure can be conditioned on additional inputs like text embeddings $\mathbf{c}_t$.
This allows to produce samples based on a text prompt.
Another control mechanism is to guide the denoising procedure with additional input maps $\mathbf{c}_\text{f}$ like segmentation masks, or landmarks.
ControlNet~\cite{controlnet} duplicates the denoising network, where one part receives the noise sample and the other the guiding conditioning. The intermediate features of the two networks are blended to achieve the conditional denoising.
Adding the conditioning on text $\mathbf{c}_t$ and images $\mathbf{c}_\text{f}$, \Cref{eq:ldmloss} is transformed to the objective:
\vspace{-3pt}\begin{equation}\vspace{-3pt}
    L = \mathbb{E}_{\mathbf{z}_0, \mathbf{t}, \mathbf{c}_t, \mathbf{c}_\text{f}, \epsilon \sim \mathcal{N}(0, 1) }\Big[ \Vert \epsilon - \epsilon_\theta(\mathbf{z}_{t}, \mathbf{t}, \mathbf{c}_t, \mathbf{c}_\text{f})) \Vert_{2}^{2}\Big] .
    \label{eq:loss}
\end{equation}
\begin{figure*}[t!]
\centering
  \includegraphics[width=0.9\textwidth]{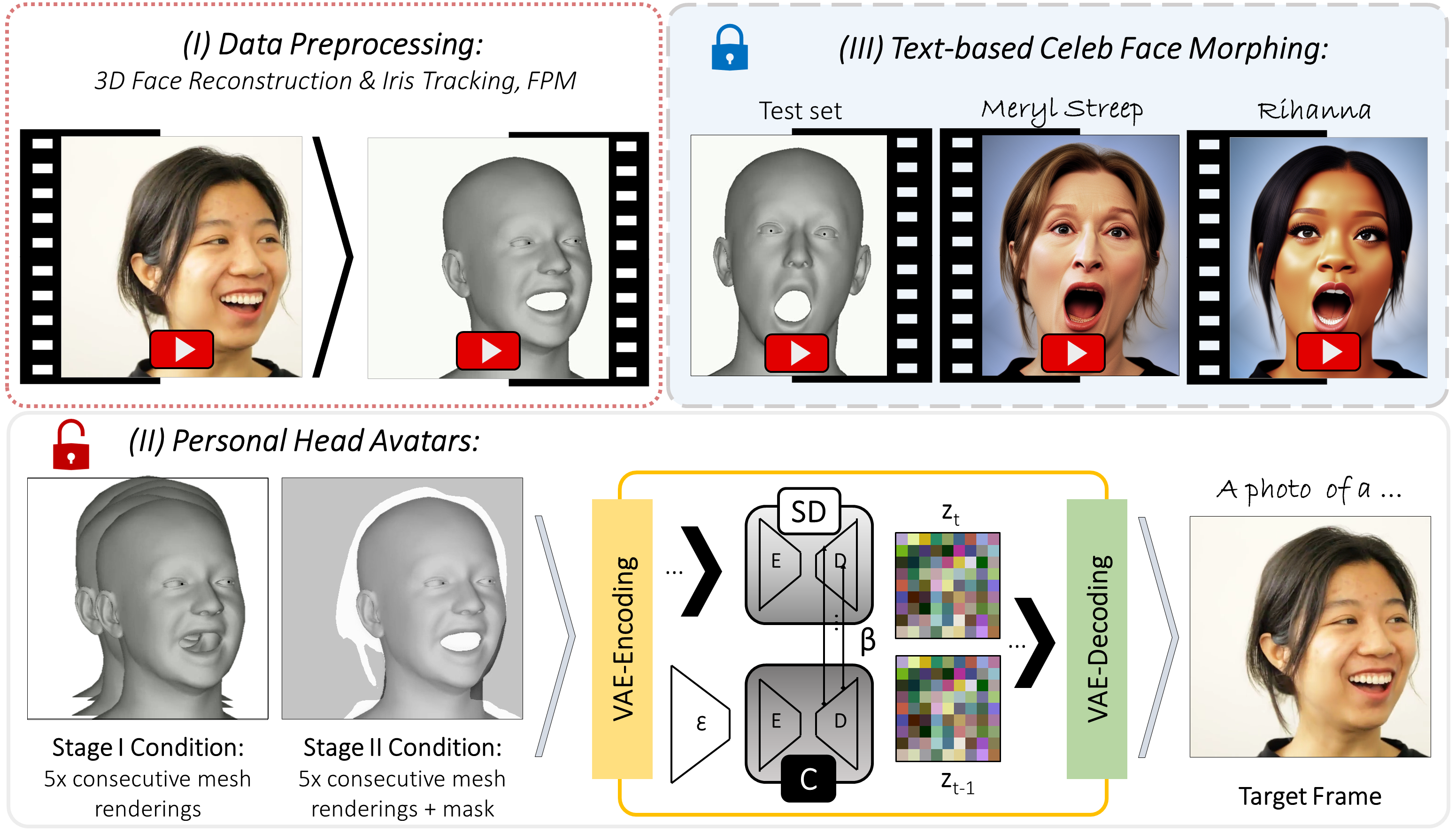}
  \caption{\textbf{System Overview:} (I) Using Spectre~\cite{spectre}, face parsing maps (FPM)~\cite{hpm}, and Mediapipe~\cite{mediapipe}, the input video is processed to extract per-frame 3D face reconstructions (3DMM), FPM, and the iris location. (II) Based on this data, two ControlNets are trained in parallel, allowing for the generation of temporally stable outlines (Stage I) and inner details (Stage II), resulting in photo-realistic personal avatars (SD is fine-tuned in the unlocked mode).
  (III) Person-specific avatars may be further morphed into a celebrity via text, without additional fine-tuning (using the locked SD).}
  \label{fig:system}
\end{figure*}
\section{Methodology}
\label{sec:method}
Based on a short video of a human subject talking, we train a personalized head avatar model leveraging the Stable Diffusion image prior (see \Cref{fig:system}).
Our method generates unseen/new videos of a subject, which is important for telepresence applications in AR/VR.
As a control, we leverage the parameter space of a 3D morphable model (3DMM) as well as textual inputs in case of appearance editing.
Our pipeline for personalized head avatar reconstruction consists of two stages to enable temporally stable outputs:
\textit{Stage I} takes in a temporal sequence of rendered 3DMM models as well as the corresponding 2D landmarks (eye pupil) as an optional input and produces a binary parsing mask for the input conditioning in 
\textit{Stage II}. The binary parsing mask is used as an additional control signal that contains temporal information on parts that are not represented by the 3DMM, e.g. the shoulders, hair, etc.
During the training of these two stages, our diffusion model is unlocked and fine-tuned simultaneously with a ControlNet module which leads to the accurate reconstruction of person-specific details.
Finally, we show how Stable Diffusion can be used to morph the person's appearance towards a text-defined celebrity.
\subsection{Monocular Head Avatars}
\label{sec:personal}
\begin{figure}[t!]
\centering
  \includegraphics[width=0.6\columnwidth]{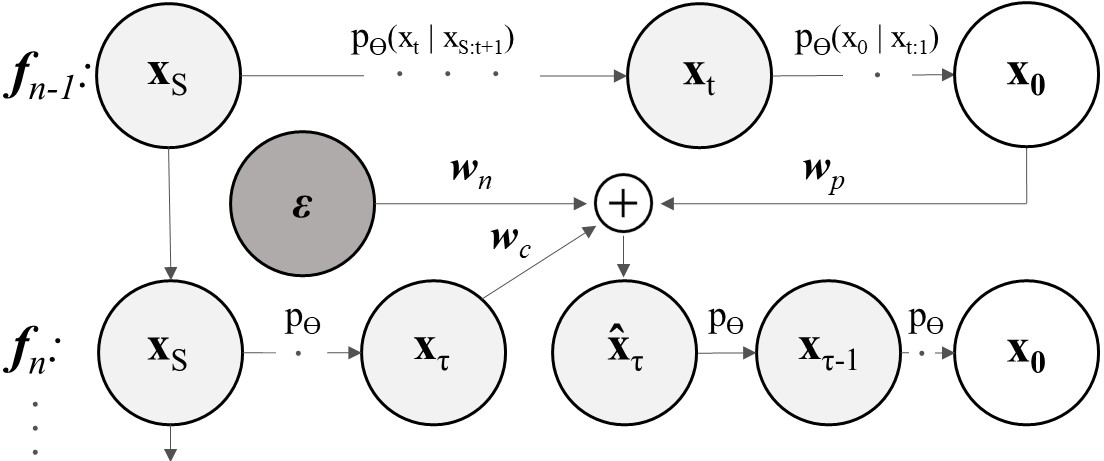}
  \caption{
      \textbf{Spatio-temporal Denoising:} Using the prediction for the frame ${f_{n-1}}$, we modify the inference in the DDIM step $t=\tau$ for frame ${f_{n}}$ to consider the previous frame, which leads to temporally smooth outputs, controlled by $w_c,w_p$ and $w_n$.
  }
  \label{fig:denoising}
\end{figure}

We take advantage of the control idea from ControlNet~\cite{controlnet} to guide the diffusion process of a latent diffusion model~\cite{stable_diffusion} conditioned on the 2D renderings of the 3DMM.
The key contribution of this work is how to handle temporal inputs and how to establish a spatio-temporal inference procedure.
As mentioned above, we employ two stages of diffusion.
Both stages rely on a modified inference procedure of the DDIM~\cite{song2020denoising} to produce temporally consistent outputs.
\paragraph{\textbf{Spatio-temporal inference procedure:}}
While LDMs~\cite{stable_diffusion} are designed to generate single images, we aim to generate temporally consistent video frames.
To this end, we modify the original inference procedure to include information about the previous frame, allowing for smoother transitions between the frames (for a graphical model, see \Cref{fig:denoising}).
At timestep $t=\tau$ ($\tau = 23$ using $S=30$ DDIM steps) of the denoising process for frame $n$, we sum the final latent prediction of the previous frame ($n-1$), the current latent prediction and a noise term $\epsilon$:
\begin{equation}
    \hat{\mathbf{x}}_{\tau}^{n} = 
        w_c \cdot \mathbf{x}_{\tau}^{n}
        + w_p \cdot \mathbf{x}_{0}^{n-1}
        + w_n \cdot \epsilon ,
    \label{eq:denoising_our}
\end{equation}
where $w_p$ is the importance of the previous frame, $w_c$ is the importance of the current frame, and $w_n$ is the amount of noise $\epsilon \sim \mathcal{N}(0, 1)$ added.
The denoising process is continued as usual after this procedure.
Adding noise $\epsilon$ in transitions where the face is moving fast helps mitigate blurriness artifacts that happen when the previous frame is far away from the current frame (see \Cref{fig:denoising-ablation}).
Note that we initialize $x_S$ with the equivalent noise term for all frames.
We use this procedure in both stages for avatar creation.
In \textit{Stage I}, we give more importance to the previous frame to generate smooth binary face parsing masks, while in \textit{Stage II}, we give more importance to the current frame.
As such, Stage I focuses on producing a temporally stable coarse outline of the avatar (hair and shoulders), while Stage II aims at smoothness and details.
\paragraph{\textbf{Training Stage I:}}
The first stage of our avatar generation method aims to generate smooth binary face parsing maps which contain a smooth outline of the person including the hair and shoulders that are not modelled by the 3DMM.
To reconstruct a 3DMM from a training sequence, we use Spectre~\cite{spectre}.
The reconstructed 3DMM model and its 2D renderings are utilized to provide the conditioning to a modified ControlNet architecture.
At test time, the 3DMM parameters can be changed and novel expressions and poses can be rendered.
The facial expression parameters estimated from Spectre and other state-of-the-art trackers~\cite{deca, deng2019accurate} might contain noise, thus, leading to inconsistent training.
We, therefore, use a series of mesh renderings as input conditioning to the ControlNet, similar to DreamPose~\cite{karras2023dreampose}.
Specifically, $5$ consecutive grayscale renderings provide a temporal conditioning signal $c_f$ (with 5 channels, resulting from the $5$ grayscale images).
To control the eyes, we additionally render the pupil~\cite{mediapipe} of the respective frame into the grayscale renderings (optional for the static eyes).
Conditioned on these controls, the ControlNet outputs an RGB image, using our proposed video inference procedure.
Note that we use a high weight $w_p$ on the previous frame to generate smooth videos, which might lack in facial details.
From generated images, human face parsing maps are estimated with~\cite{hpm}.
We train the first stage ControlNet model in the unlocked mode, where both LDMs are fine-tuned simultaneously to match the appearance of the subject as closely as possible, lowering the possiblity of failure given that we take advantage of a pre-trained face parsing model on top of our output.
We use \Cref{eq:loss} as a training objective, where $c_f$ is our custom control image encoded into a feature space conditioning vector through a small neural network~\cite{controlnet}.
\paragraph{\textbf{Training Stage II:}}
The second stage is used to generate the details that are missing in the temporally smooth outputs of the first stage.
Specifically, smooth human face parsing maps generated from the first stage are input to the second diffusion model.
During training of this second ControlNet, parsing maps estimated from the original training samples~\cite{hpm} are used, thus, the two stages can be trained in parallel, as the output of the first stage is not required during training of the second.
The model of the second stage receives three channels as input: (I) the binary parsing map, (II) the grayscale mesh rendering of the current frame with iris renderings for eye control, as well as (III) a motion map.
The motion map for frame $n$ is computed by accumulating the renderings of the meshes $n-2,n-1, n+1$, and $n+2$,
with weights $\frac{1}{6}, \frac{1}{3}, \frac{1}{3}, \frac{1}{6}$, respectively.
The model is trained with the same scheme as in \textit{Stage I}, however, the inference weights for the previous frame are set to a lower value (see \Cref{fig:denoising-ablation}).
\subsection{Text-based Celeb Face Morphing}
\begin{figure*}[t!]
\centering
    \includegraphics[width=1.0\textwidth]{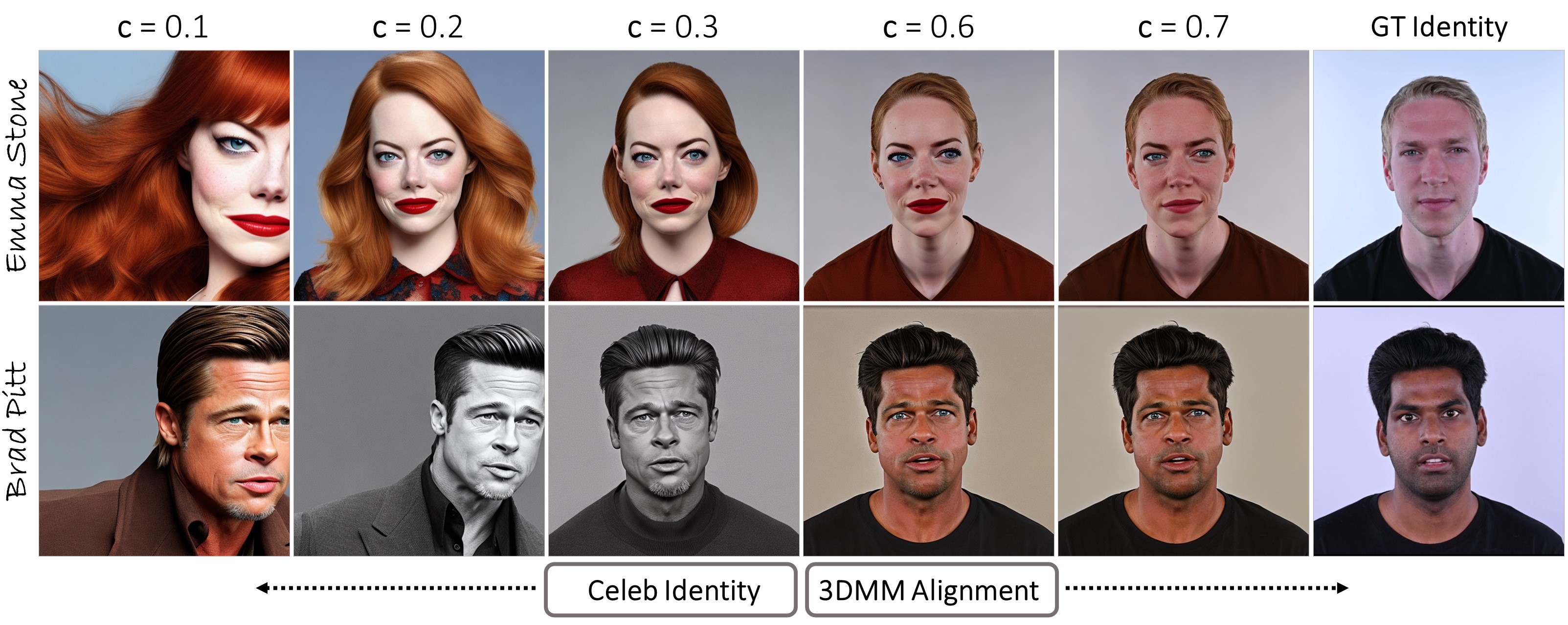}
  \caption{\textbf{ControlNet Strength:} The lower values give more importance to the celeb ID (as defined by text), but they lead to inconsistent videos and low controllability via 3DMM. The higher values allow for more control while preserving the original identity.}
  \label{fig:ablation-control-stren}
\end{figure*}

While the Stable Diffusion model for the personalized avatar creation is unlocked during training and adapts to the idiosyncrasies of a subject, to enable morphing via text, we do not fine-tune it in stage II of our text-based generative facial celebrity appearance morphing model.
An important factor during inference with this model is the ControlNet strength.
The ControlNet strength $c$ weights the features from the pre-trained Stable Diffusion part and the fine-tuned ControlNet module (see \Cref{fig:ablation-control-stren}).
A weight of $c=1$ signifies that the personal ControlNet model has full impact while the Stable Diffusion model has a low influence, disabling the usage of text.
Vice versa, with a weight of $c=0$, the pose and expression information of the ControlNet is ignored, and an image is generated that follows the text prompt.
We found that $c<0.5$ results in temporally unstable results that are not well controlled by the 3DMM.
In addition to the ControlNet strength parameter $c$, the classifier-free guidance scale of the pre-trained Stable Diffusion network controls the resulting appearance.
Higher values lead to an output that matches more closely to the appearance specified by the text prompt, however, when the weight is too high, it may produce results of low quality that resemble caricatures and do not look like real humans.
In our experiments, we use a default classifier guidance scale factor of $7.5$.
While the method described above gives stable content for the human (foreground), the background might be temporally unstable.
To this end, we employ an additional background inpainting using the predicted masks and the procedure described in Sec. 4.5 of the original Stable Diffusion work~\cite{stable_diffusion}.
\begin{figure*}[ht!]
\centering
    \includegraphics[width=\textwidth]{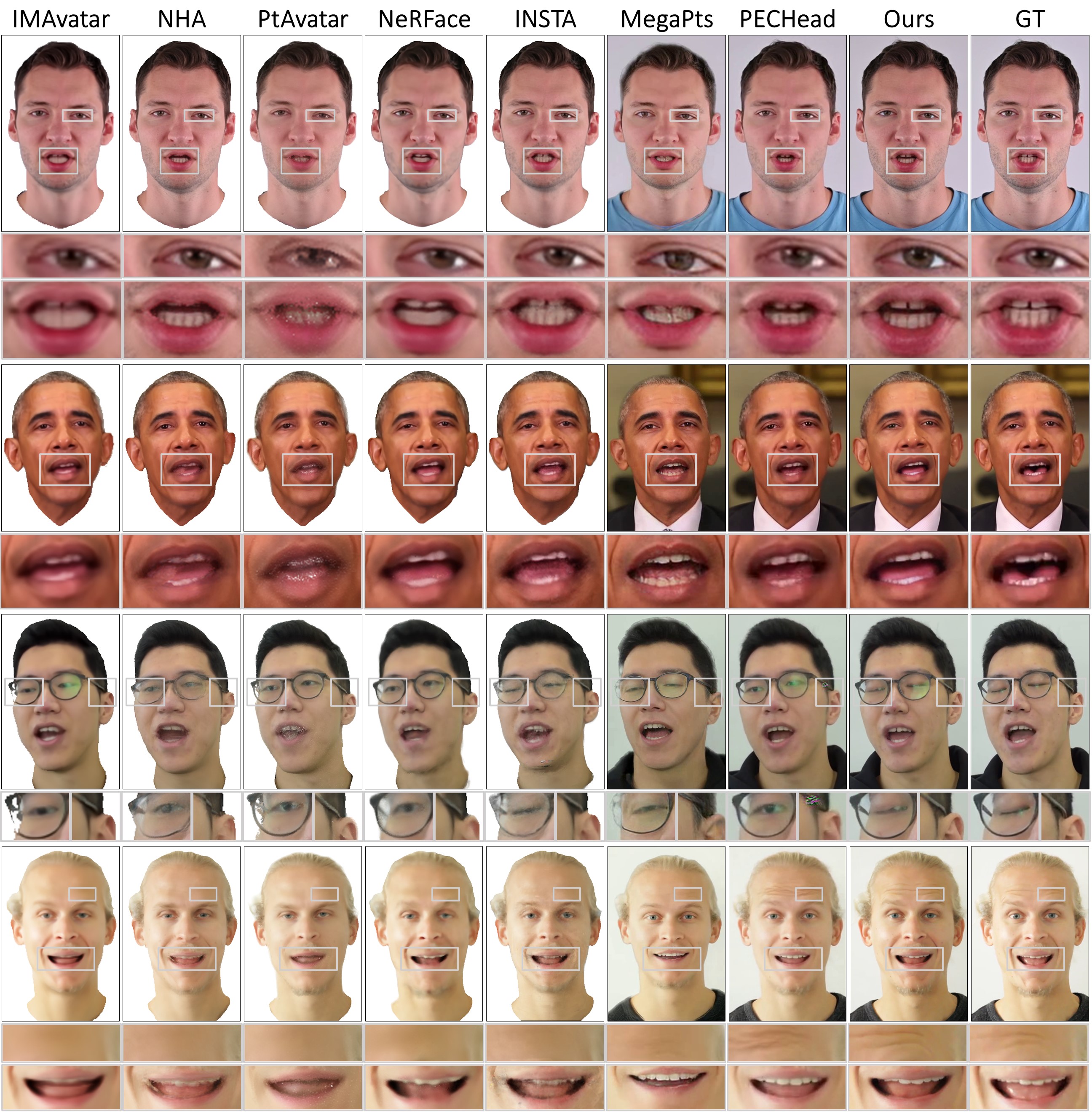} 
  \caption{\textbf{Qualitative Comparison with State of the Art:} Our method produces more detailed results in cases where features such as wrinkles (row 4), mouth interior (rows 2, 4), teeth (rows 1,--4), hair/beard (rows 1, 4), or eyeglasses (row 3) appear.}
  \label{fig:final_results}
\end{figure*}

%% file: sec/4_experiments.tex
\begin{figure}[t!]
\centering
    \includegraphics[width=1.0\columnwidth]{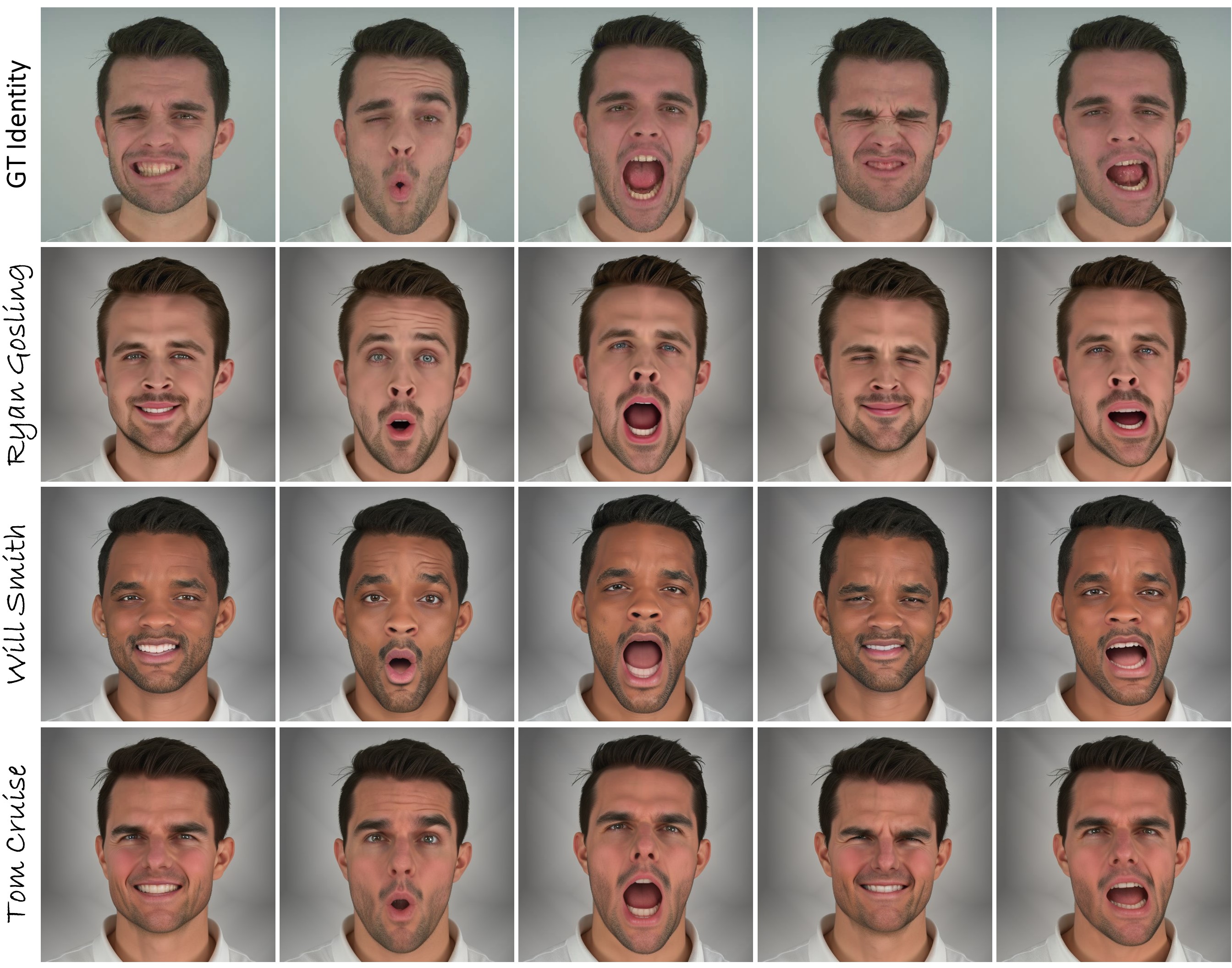}
    
  \caption{\textbf{Face Morphing:}
  Our method implicitly morphs the person's appearance (row 1) with a celebrity (rows 2-4).
  Identity remains stable under challenging expressions.
  }
  \label{fig:id-male}
\end{figure}
\section{Experiments}
\label{sec:experiments}
To evaluate \textit{SVP}, we run experiments on single-view data used in state-of-the-art avatar reconstruction methods as well as on multi-view data from Nersemble~\cite{nersemble}.
The single-view data is used to compare our portrait avatar generations to existing methods, while the multi-view data (see \Cref{fig:teaser}) is treated as a monocular sequence to demonstrate the stability of our celeb face morphing approach.
For more qualitative results, please see our supplementary materials.

\subsection{Monocular Personalized Head Avatars}

\begin{table}[tb]
\caption{\textbf{Data Reduction Study:} We evaluate the impact of reducing the amount of training data by keeping only a specific percentage of frames (see the \% column).}
\centering
\begin{tabular}{ccccccccc}
    \toprule
     & \multicolumn{4}{c}{\textit{Sequence I: \textbf{Obama}}} & \multicolumn{4}{c}{\textit{Sequence II: \textbf{Biden}}} \\
    \cmidrule(lr){2-5} \cmidrule(lr){6-9}
    \textbf{ \% } & PSNR $\uparrow$ & SSIM $\uparrow$ & MSE $\downarrow$ & LPIPS $\downarrow$ 
    & PSNR $\uparrow$ & SSIM $\uparrow$ & MSE $\downarrow$ & LPIPS $\downarrow$ \\
    \midrule
    \textbf{100} & 33.78 & 0.954 & 0.0004 & 0.016 & 34.41 & 0.960 & 0.0004 & 0.015 \\
    \textbf{50} & 32.59 & 0.951 & 0.0006 & 0.016 & 32.56 & 0.955 & 0.0006 & 0.017 \\
    \textbf{25} & 30.94 & 0.939 & 0.0008 & 0.021 & 29.69 & 0.944 & 0.0011 & 0.023 \\
    \textbf{12.5} & 26.49 & 0.916 & 0.0023 & 0.035 & 26.46 & 0.929 & 0.0023 & 0.032 \\
    \textbf{6.25} & 20.99 & 0.864 & 0.0080 & 0.132 & 20.72 & 0.882 & 0.0086 & 0.107 \\
    \bottomrule
\end{tabular}
\label{tab:obama_biden_metrics}
\end{table}

\paragraph{\textbf{Quantitative evaluation:}}
In \Cref{tab:sota_comparison}, a quantitative comparison against state-of-the-art monocular avatar reconstruction methods is shown.
For this comparison, we use monocular videos from the publicly released datasets of INSTA~\cite{zielonka2023instant}, IMAvatar~\cite{zheng2022avatar}, NHA~\cite{grassal2022neural}, as well as of PointAvatar~\cite{zheng2023pointavatar}.
As baselines we consider IMAvatar~\cite{zheng2022avatar}, NHA~\cite{grassal2022neural}, PointAvatar~\cite{zheng2023pointavatar}, NeRFace~\cite{gafni2021dynamic}, INSTA~\cite{zielonka2023instant}, MegaPortraits~\cite{drobyshev2022megaportraits}, and PECHead~\cite{gao2023high}.
All methods, except, MegaPortraits which is a one-shot method, are trained on the entire training video (per subject).
Our method generates high-quality images which is reflected in the image reconstruction metrics in \Cref{tab:sota_comparison}.
Note that the metrics are computed using the scheme of INSTA, where only the head and neck regions are considered, as most of the methods only reconstruct those regions.
Additionally, \Cref{tab:obama_biden_metrics} provides a quantitative analysis considering the amount of data used for training in two separate sequences, featuring Biden and Obama.
As expected, our method performs best with the most data, while performance drops as data is reduced.
This highlights the importance of using sufficient training data for optimal results.
\paragraph{\textbf{Qualitative evaluation:}}
The qualitative results on the task of personalized avatar reconstruction are shown in \Cref{fig:final_results}.
Our qualitative analysis reveals that our method excels in specific areas, outperforming the baseline methods when facial features such as wrinkles (row 4), the interior of the mouth (rows 2 and 4), teeth (rows 1, 2, 3, 4), hair and beard (rows 1, 4), or eyeglasses (row 3) appear.
\subsection{Celeb Face Morphing}
Our method allows us to morph the face considering the identity, expressions and head pose, as shown in \Cref{fig:teaser}.
In \Cref{fig:id-male}, we show a further example of identity changes based on monocular data.
The key finding is that our method generates faces that may be morphed with celeb faces, without additional fine-tuning, and these identity morphs remain stable throughout the video even for challenging expressions, provided that similar expressions exist in training data.

\input{tab/tab_sota_comparison_1col.tex}
\subsection{Ablation Studies}
\begin{figure}[t!]
\centering
    \includegraphics[width=0.9\columnwidth]{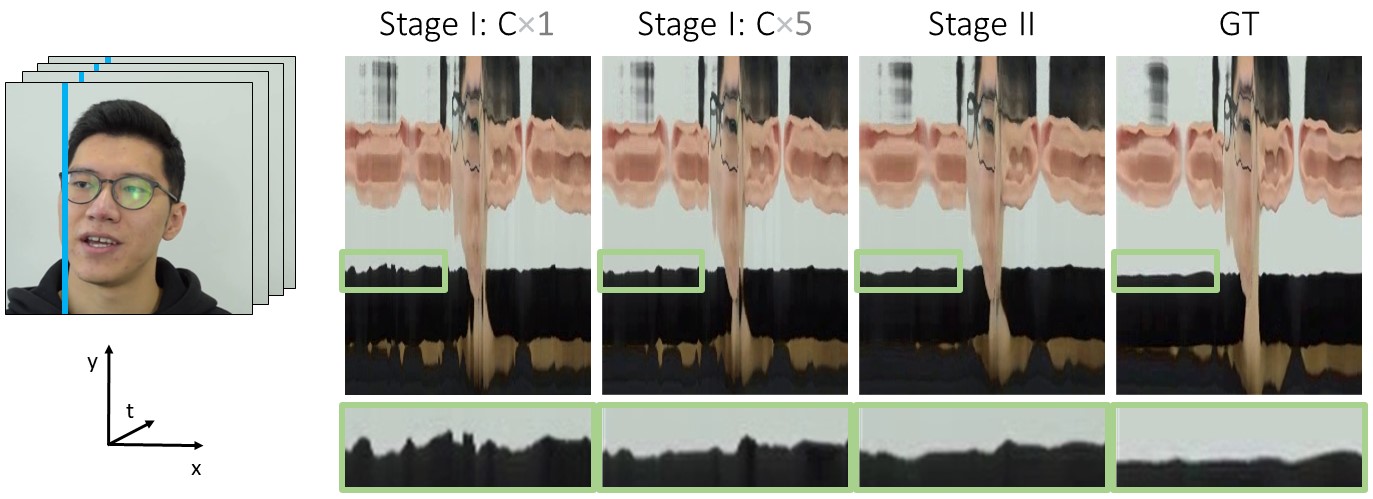}
  \caption{\textbf{Ablation on Temporal Coherence:} Our stage II prediction leads to smoother transitions as compared to the stage I prediction using either single or chained controls (1x or 5x meshes) as input conditioning. The improvement can be observed especially in the areas that are not controlled by a 3DMM. Here, the shoulders region is showcased. We show a video extract corresponding to the movements of one vertical stripe (on the left, in blue) over time. Please see the supplemental video for details.}
  \label{fig:volume}
\end{figure}

\begin{figure}[t!]
\centering
    \includegraphics[width=0.9\columnwidth]{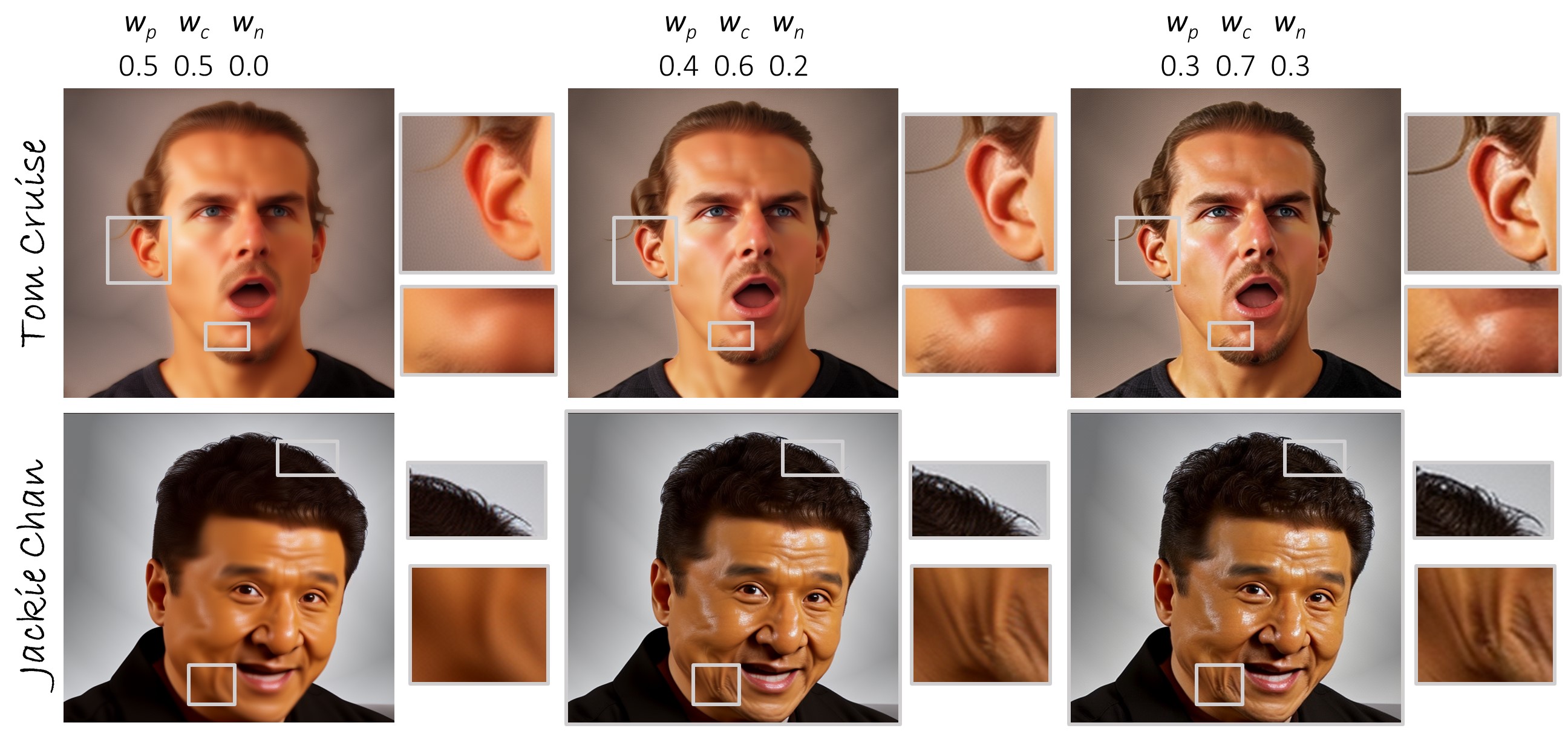}
  \caption{\textbf{Ablation of the Denoising Parameters:} A high impact of the previous frame (colum 1) leads to blurry but temporally smooth results. Adding random noise (columns 2, 3) helps in generating details like facial hair, but it might generate images that are too sharp (column 3). With reduced influence of the previous frame, temporal consistency is lowered and flickering artifacts appear. Please see the supplemental video.}
  \label{fig:denoising-ablation}
\end{figure}

\paragraph{\textbf{Denoising process:}}
The spatio-temporal denoising process introduces additional parameters that allow for a control over how much influence the previous frame has on the current frame.
In the suppl. doc., we provide a detailed quantitative analysis of these parameters.
We visualize the effect of different parameters used in this study in \Cref{fig:denoising-ablation}.
High influence from the previous frame causes blurriness, while added noise enhances details like facial hair.
Dropping the information of the previous frame (as in the original denoising scheme) leads to temporally unstable results (\Cref{fig:volume}).
\begin{wrapfigure}[28]{t!}{0.57\columnwidth}
  \centering
  \includegraphics[width=0.57\columnwidth]{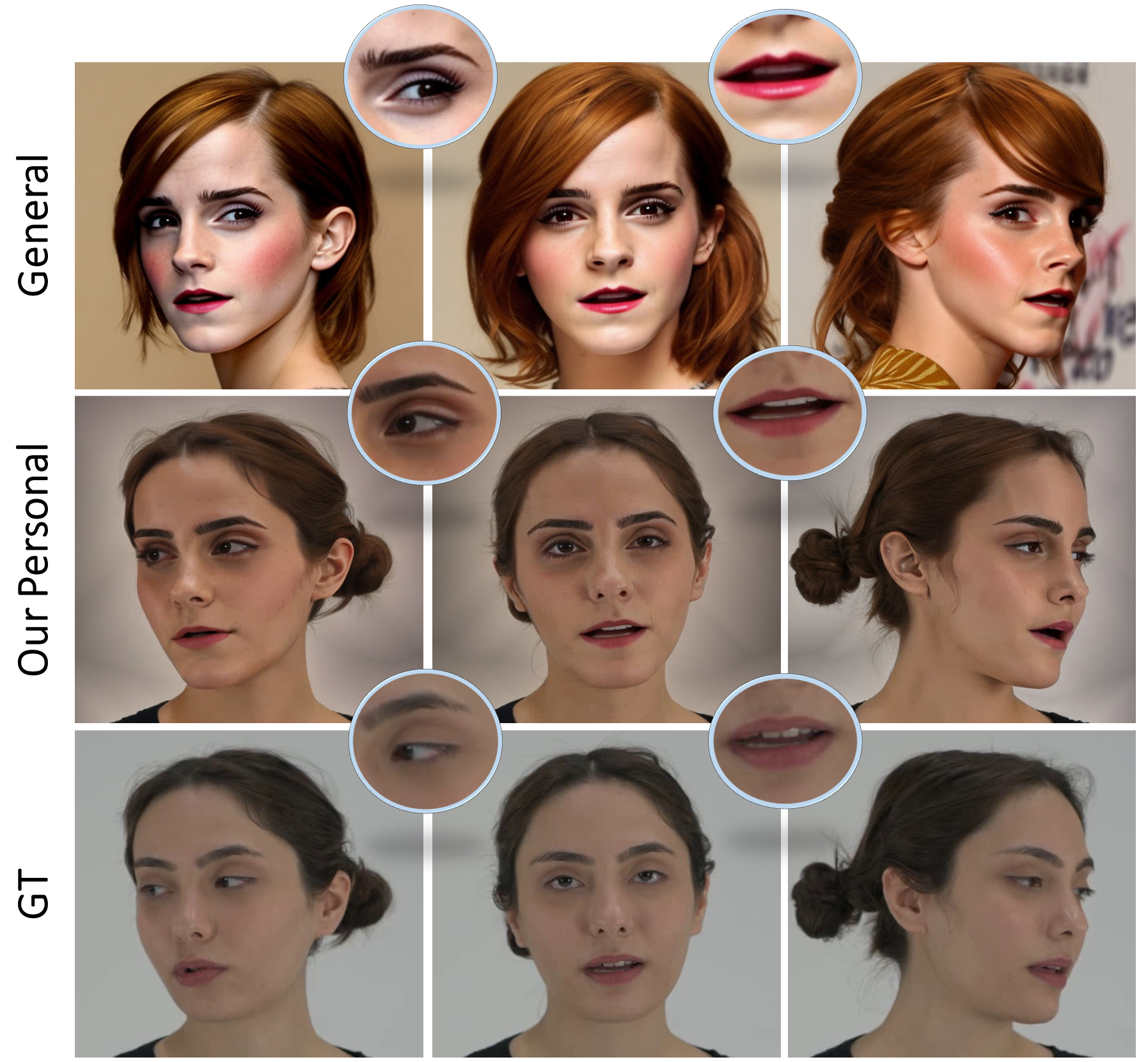}
  \caption{\textbf{Person-Specific vs. Person-Generic Training:} Generating a video of a person defined by text and a sequence of facial animation parameters (3DMM) is a challenging task for person-generic latent diffusion models (row 1, general). This difficulty arises due to the high ambiguity in mapping between the low-frequency conditioning and the high-dimensional latent space. To address this, we propose fine-tuning Stable Diffusion on one personal video sequence (row 3, GT) to achieve stable video generation. The generated faces can be further morphed with a celebrity, as shown in row 2 (personal) with the text prompt \textit{"Emma Watson"}, while maintaining stability.}
  \label{fig:sd-baseline}
\end{wrapfigure}
\paragraph{\textbf{Person-specific v.s. person-generic model:}}
In \Cref{fig:sd-baseline}, we give a comparison between a personalized and a general model.
For this comparison, we fine-tune Stable Diffusion using a ControlNet module on data from FFHQ and CelebA-HQ that contains a total of 100K images, using 3DMM renderings as conditioning.
As can be seen, the person-generic model generates temporally inconsistent identity images.
In contrast, our method produces consistent appearances under challenging expressions as well as views (see \Cref{fig:teaser}).
\paragraph{\textbf{ControlNet strength:}}
\Cref{fig:ablation-control-stren} shows the impact of the ControlNet strength parameter $c$.
Low guidance from the ControlNet results in image generations that ignore pose and expression information from the 3DMM.
By varying the strength parameter, different blending results can be achieved.

\paragraph{\textbf{Limitations:}}
Our approach comes with the following downfalls.
3D face reconstruction errors may result in lowered faithfulness to the original expressions (see \Cref{fig:id-male}) or image quality might suffer, especially, if the provided test parameters are corrupted.
The 3DMM also does not provide information about the tongue, and dynamics of the hair.
Supplying sufficiently varied training data will lead to better expression generalization.

%% file: tab/tab_sota_comparison_1col.tex
\begin{table}[tb]
\caption{\textbf{Quantitative Comparison with State-of-the-art:} Cells contain the mean error (columns) for each method (rows) and for 8 sequences (from INSTA (4), NeRFace (2), NHA (1), and IMAvatar (1)).
Only the head + neck region is considered.
Our method is of high fidelity and it is competitive with SOTA in PSNR, SSIM, and MSE metrics, while it outperforms SOTA in LPIPS, FID, and KID metrics.
}
\centering
    \begin{tabular}{ccccccc}
    \toprule
     Method  & PSNR $\uparrow$& SSIM $\uparrow$ & MSE $\downarrow$ & LPIPS $\downarrow$ & FID $\downarrow$ & KID $\downarrow$\\
    \midrule
    I'M Avatar~\cite{zheng2022avatar} & 26.0  & 0.9363  & 0.0030  & 0.079 & 48 & 0.062  \\
    MegaPortraits~\cite{drobyshev2022megaportraits} &  27.7  & 0.9085 & 0.0021  & 0.057 & 37  & 0.041  \\
    NHA~\cite{grassal2022neural}  & 28.1  & 0.9417  & 0.0019  & 0.050  & 25  & 0.023  \\
    Point Avatar~\cite{zheng2023pointavatar} & 28.6  & 0.9301 & 0.0015  & 0.063  & 33  & 0.037  \\
    NeRFace~\cite{gafni2021dynamic}& 29.0  & \textbf{0.9487} & 0.0016  & 0.063  & 44  & 0.056  \\
    INSTA~\cite{zielonka2023instant}  & 28.2  & 0.9397  & 0.0017  & 0.061  & 31  & 0.033  \\
    PECHead~\cite{gao2023high}  & \textbf{31.4} & 0.9459  & \textbf{0.0009} & 0.034  & 23 & 0.025  \\
    \textbf{Ours}&  \textbf{31.4}  & 0.9445 &0.0010  & \textbf{0.026 } & \textbf{11} & \textbf{0.005 }\\
    \bottomrule
    \end{tabular}
\label{tab:sota_comparison}
\end{table}

%% file: sec/5_conclusion.tex
\section{Conclusion}
\label{sec:conclusion}
In this paper, we have introduced \textit{Stable Video Portraits}, a method for generating controllable and morphable 3D avatars.
Specifically, we leverage a Stable Diffusion prior through ControlNet, guided by a temporal 3DMM sequence.
Our novel denoising scheme leads to the generation of temporally stable videos.
Compared to current monocular avatar reconstruction methods, it achieves state-of-the-art quality, with the additional possibility of face morphing with celeb faces.

%% file: sec/X_suppl.tex
\clearpage

\title{Stable Video Portraits - Appendix} 

\author{Mirela Ostrek\inst{1}\orcidlink{0009-0009-9987-646X} \and
Justus Thies\inst{1,2}\orcidlink{0000-0002-0056-9825}}

\authorrunning{M. Ostrek et al.}

\institute{Max Planck Institute for Intelligent Systems, Tübingen, Germany \and
Technical University of Darmstadt, Darmstadt, Germany}

\maketitle

\setcounter{page}{1}

\begin{figure}
\centering
  \includegraphics[width=0.8\textwidth]{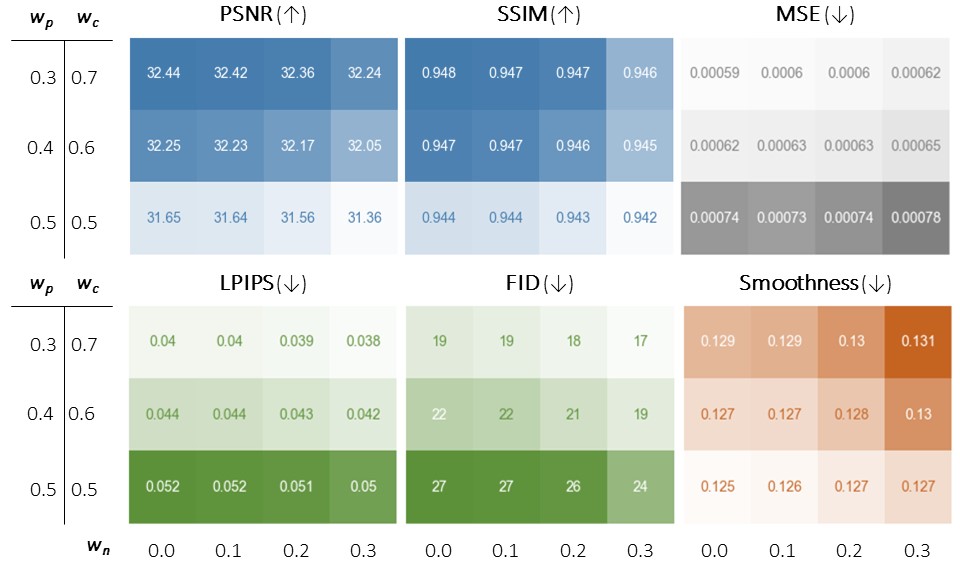}
  \caption{\textbf{Quantitative ablation study:} To investigate the effect of each of the newly introduced denoising process parameters, namely $w_n$ (noise term importance), $w_c$ (importance of the current frame), and $w_p$ (importance of the previous frame), we show the results on five standard evaluation metrics including PSNR, SSIM, MSE, LPIPS, FID including our proposed smoothness metric. Darker cells contain higher values.}
  \label{fig:cap}
\end{figure}
\section{Training Details}
For avatar reconstruction, we fine-tune Stable Diffusion 2.1 model on each sequence using ControlNet.
Our input control is comprised of 5 consecutive meshes with two eye landmarks rendered on top of each, and we additionally use alpha masks that cover the head and neck region (provided in INSTA).
We train the model for 200 epochs, with batch size 16, and gradient accumulation set to 4.
The following text prompt is used and dropped 50\% of the time for each sequence during training: "a high-quality, detailed, and professional photograph, portrait of a person, speaking".
At test time, we set the CGS to 5 and we do not use additional text prompts.
We randomly pick a number as a random seed and we keep it fixed for every sequence during the entire generation process.
DDIM is used and the number of sampling steps is set to 30.

\section{Data}
A portrait avatar dataset has been released, containing six long speaking sequences, each lasting more than eight minutes. These sequences include head movement and feature female subjects with fixed hair.
In \Cref{fig:data}, the samples are shown.
Please see the supplemental video for celebrity reconstructions featuring some of the subjects.

\begin{figure}[b!]
\centering
  \includegraphics[width=1.0\textwidth]{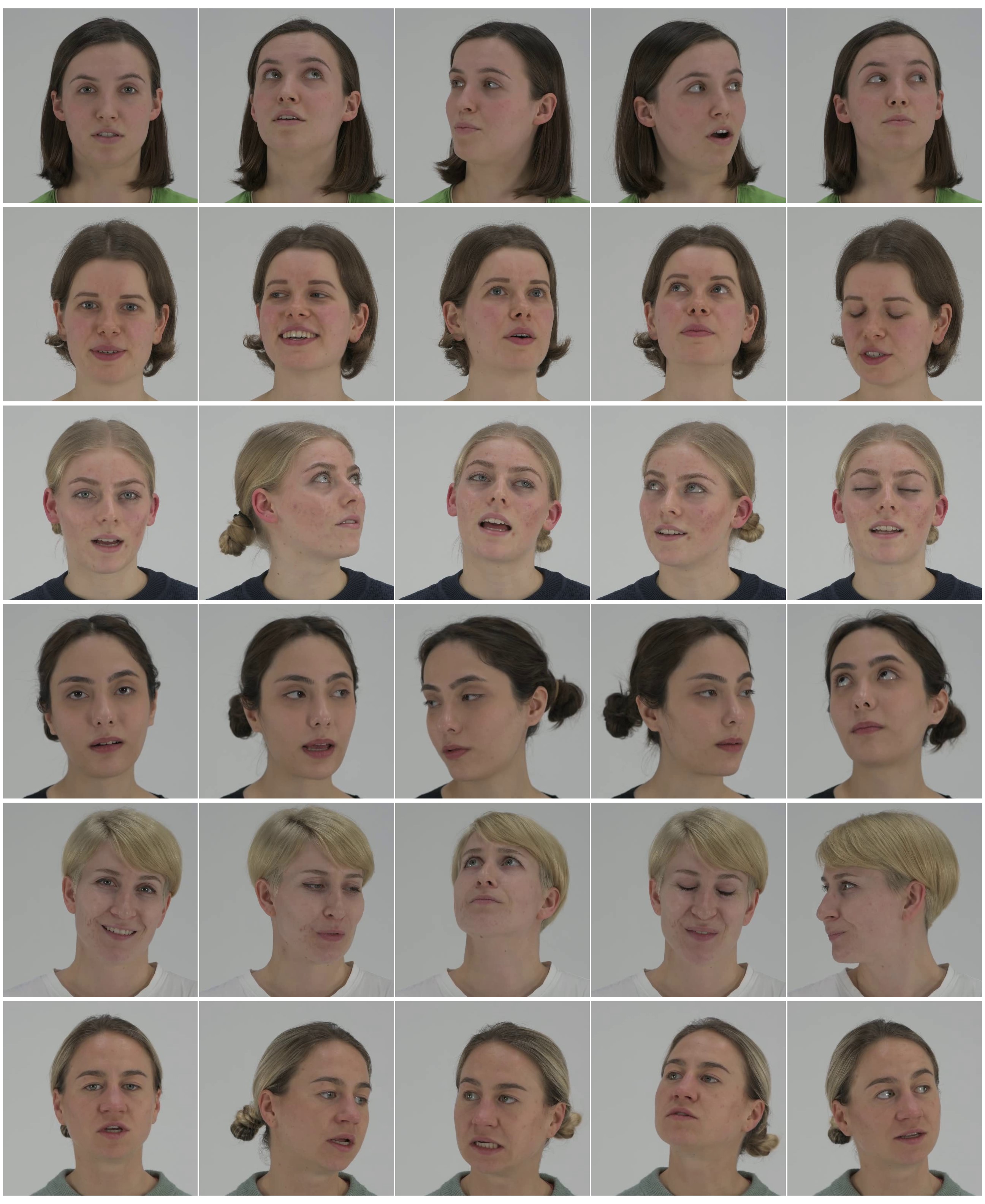}
  \caption{\textbf{Data:} We have released \textit{a portrait avatar dataset} that contains 6 long video sequences ($8+$ minutes) of women speaking with head movement, for research purposes. }
  \label{fig:data}
  \vspace{-20pt}
\end{figure}
\begin{figure}[t!]
\centering
  \includegraphics[width=0.9\textwidth]{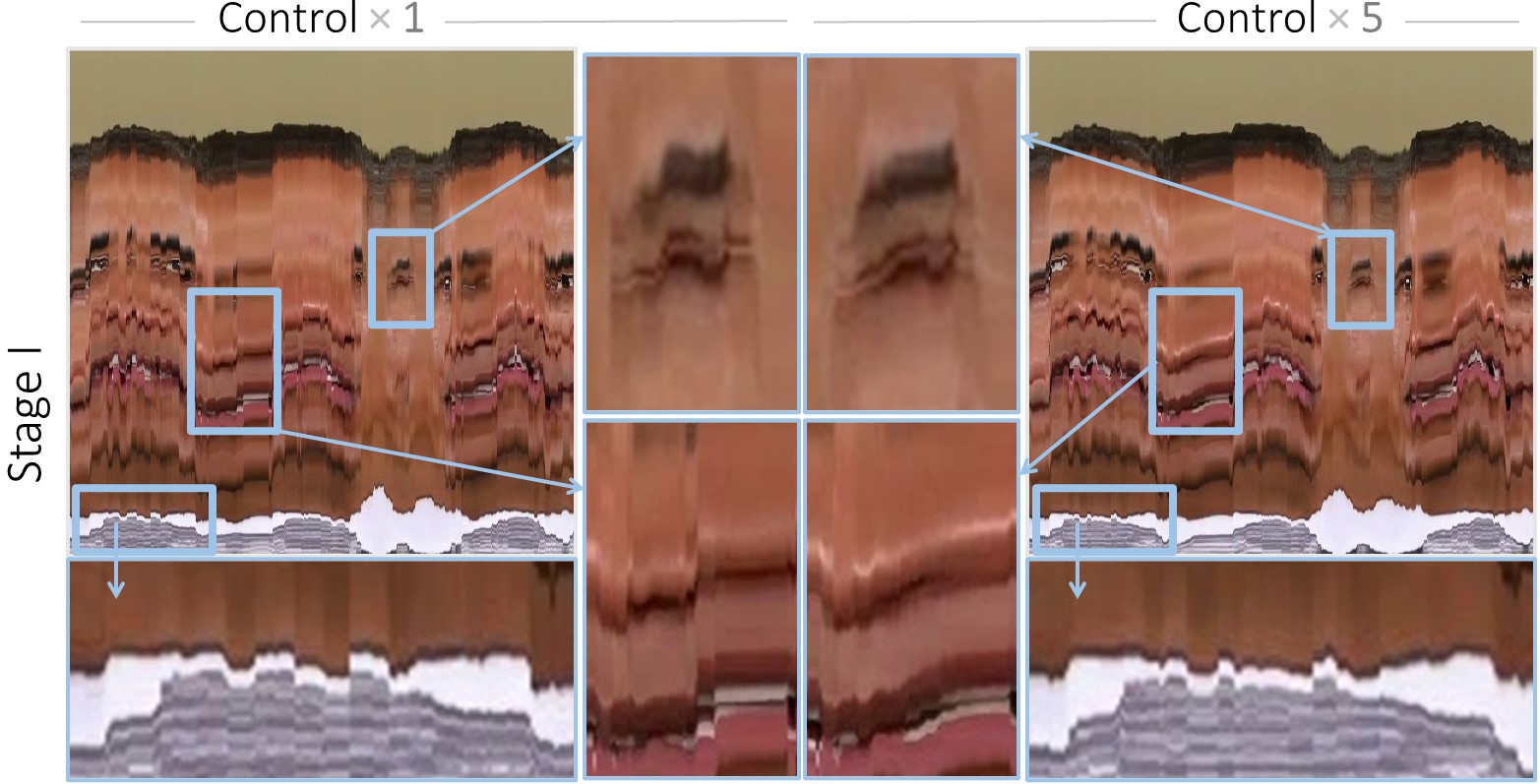}
  \caption{\textbf{Temporal conditioning:} Using temporal control signal reduces temporal incoherence due to noisy conditioning. Here, we show how one column of the generated image changes over time when only 1 v.s. 5 consecutive 3DMM controls are used.
  }
  \label{fig:ablation-input}
\end{figure}
\section{Additional Experiments}
\label{sec:add_experiments}

\subsection{Quantitative Ablation Study on the Denoising Parameters}
In Figure~\ref{fig:cap}, we investigate the properties of each of the newly introduced parameters of the denoising process, considering five standard evaluation metrics (PSNR, SSIM, MSE, LPIPS, FID) and we propose an evaluation metric for smoothness, i.e. how smooth the overall transitions in the generated videos are.
The latter is defined as a sum of the MSE error between the two neighboring frames. 
The lower value indicates a smoother transition.
All of the values are calculated for one example sequence.
It can be seen how giving more importance to the previous frame through $w_p$ leads to smoother transitions (last matrix).
However, adding noise through $w_n$ makes the smoothness error higher.
At the same time, LPIPS and FID scores consistently improve when more noise gets added as well as when $w_p$ is decreased.
Standard error metrics that encourage blurriness (MSE, PNSR, SSIM) do better when $w_n$ is equal to 0 and when $w_p$ is reduced.
We conclude that there are trade-offs when different metrics are considered.
And therefore, the most suitable set of parameters is to be determined subjectively upon visual inspection.

\subsection{Qualitative Ablation Study on Temporal Input Controls}
Figure~\ref{fig:ablation-input} contains an ablation study on the importance of using temporal controls.
Here we show that using 5 consecutive 3DMM control signals as opposed to 1 leads to the smoother transitions between the frames.

\subsection{User Study on Face Morphing}
\begin{figure}[t!]
\centering
  \includegraphics[width=\textwidth]{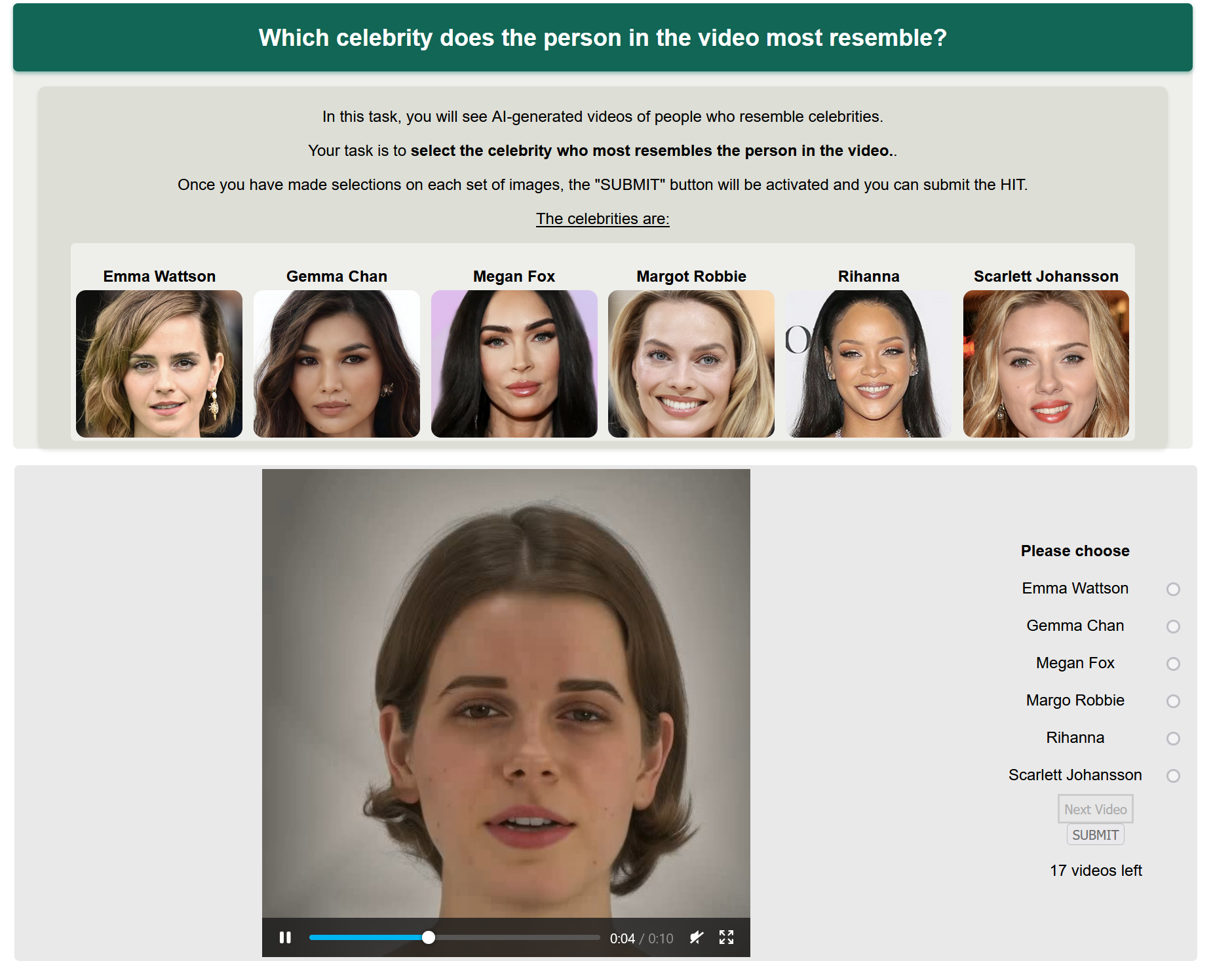}
  \caption{\textbf{User study:} Layout of our user study.
  }
  \label{fig:user}
\end{figure}

We conducted a preliminary user study involving 32 participants who viewed our celebrity face-morphing videos and were tasked with identifying the celebrity the video most closely resembled (refer to Fig.~\ref{fig:user} for more details).
The study involved 2 subjects, each morphed with 6 celebrities.
The results indicate that participants correctly identified the celebrity 50.78\% of the time, while in 49.22\% of cases, they selected one of the 5 incorrect celebrities.
